\pdfoutput=1
\documentclass[11pt]{article}

\usepackage[final]{acl}

\usepackage{times}
\usepackage{latexsym}

\usepackage[T1]{fontenc}
\usepackage[utf8]{inputenc}

\usepackage{microtype}

\usepackage{inconsolata}

\usepackage{graphicx}

\usepackage{amsmath,nccmath,tabularx} %amssymb
\usepackage{algorithm}
\usepackage{algorithmic}
\usepackage{color}
\usepackage{graphicx}
\graphicspath{{images/}}
\usepackage{wrapfig,lipsum}
\usepackage{tabularx}
\usepackage{graphicx}
\usepackage{lscape}
\usepackage{subcaption}
\usepackage{latexsym}
\usepackage{pifont}% http://ctan.org/pkg/pifont
\usepackage{multirow}

\usepackage{enumitem}
\setlist{leftmargin=6mm}
\usepackage[T1]{fontenc}
\usepackage[utf8]{inputenc}
\usepackage[font=small,labelfont=bf]{caption}

\usepackage{booktabs}

\usepackage{xcolor}
\usepackage{xspace}
\usepackage{bm}
\usepackage{amsmath}
\usepackage{placeins}

\usepackage{balance}

\newcommand{\eg}{\hbox{\emph{e.g.,}}\xspace}
\newcommand{\ie}{\hbox{\emph{i.e.,}}\xspace}
\newcommand{\ours}{\textsc{E\textsuperscript{2}R-FLOPs}\xspace}
\newcommand{\metric}{Efficiency-Effectiveness Reranking FLOPs\xspace}

\title{\metric for LLM-based Rerankers}

\author{
 \textbf{Zhiyuan Peng\thanks{Correspondence: \texttt{zpeng@scu.edu}}\textsuperscript{1}},
 \textbf{Ting-Ruen Wei\textsuperscript{1}},
 \textbf{Tingyu Song\textsuperscript{2}},
 \textbf{Yilun Zhao\textsuperscript{3}}
\\
 \textsuperscript{1}Santa Clara University,
 \textsuperscript{2}Independent Researcher,
 \textsuperscript{3}Yale University
}

\begin{document}
\maketitle
\begin{abstract}
% LLM-based rerankers enhance key ranking metrics, such as NDCG and MRR, but are often too computationally intensive for deployment. 
Large Language Models (LLMs) have recently been applied to reranking tasks in information retrieval, achieving strong performance. However, their high computational demands often hinder practical deployment.
Existing studies evaluate the efficiency of LLM-based rerankers using proxy metrics such as latency, the number of forward passes, input tokens, and output tokens. However, these metrics depend on hardware and running-time choices (\eg parallel or not, batch size, etc), and often fail to account for model size, making it difficult to interpret and obscuring the evaluation of the efficiency-effectiveness tradeoff. 
To address this issue, we propose \ours\footnote{\url{https://github.com/zhiyuanpeng/EER-FLOPs}.} for LLM-based rerankers: RPP (ranking metrics per PetaFLOP), measuring how much ranking quality (e.g., NDCG or MRR) a method achieves per PetaFLOP, and QPP (queries per PetaFLOP), measuring how many queries can be processed per PetaFLOP. Accompanied by the new metrics, an interpretable FLOPs estimator is developed to estimate the FLOPs of an LLM-based reranker even without running any experiments. Based on the proposed metrics, we conduct comprehensive experiments to evaluate a wide range of LLM-based rerankers with different architectures, studying the efficiency-effectiveness trade-off and bringing this issue to the attention of the research community. 
\end{abstract}

\section{Introduction}

A typical search system balances efficiency and quality with a two-stage pipeline: a lightweight retriever retrieves hundreds of documents from a vast corpus, prioritizing efficiency, and then a more powerful but computationally expensive reranker refines their order. 
Thanks to the rapid progress of LLMs~\cite{brown2020language, grattafiori2024llama, DBLP:journals/corr/abs-2312-11805}, LLM-based rerankers have achieved impressive gains in reranking metrics, such as NDCG; however, these gains often come at the cost of substantial computational expense, making them difficult to deploy at scale in production. This underscores the need for evaluation metrics that consider not only reranking quality but also computational efficiency.

Existing studies evaluate the efficiency of LLM-based rerankers using proxies such as latency~\cite{DBLP:conf/www/JinPZCZXLFZDM25}, the number of LLM calls (\ie forward passes)~\cite{DBLP:conf/sigir/ZhuangZKZ24}, and input and output token usage~\cite{DBLP:conf/www/0004LZ0MYSMY25}. 
% Each of these metrics has significant limitations. 
However, these metrics lack the computational granularity needed to distinguish differences in internal compute per token or per model call. 
Specifically, latency is heavily dependent on hardware and runtime choices (GPU vs. CPU, batch size, parallelism), making it an inconsistent basis for comparing algorithms across studies. The number of LLM calls ignores the model size: a single call to a 70B LLM costs orders of magnitude more compute than a call to a 3B model, yet both appear identical under this metric. Similarly, token usage overlooks the model size and is difficult to interpret as the cost of the input token and the output token can be different. 
% Consider the following scenario of two LLM-based rerankers: one with more input tokens but fewer output tokens, and the other with fewer input tokens but more output tokens. It is challenging to decide which one is more efficient, especially when the two rerankers have the same total number of input and output tokens. 
% In other words, prior efficiency metrics lack computational granularity to capture differences in the internal compute per token or per call.

Inspired by the scaling law in LLMs that studies the connection between total compute and performance ~\cite{DBLP:journals/corr/abs-2001-08361}, we employ floating-point operations (FLOPs) as a fundamental measure of cost for each forward pass or LLM call. The total number of FLOPs required by a model to rerank documents is a hardware-agnostic, intrinsic metric of computational work ~\cite{DBLP:conf/nips/SukthankerZSKPF24}. Building on this insight, we introduce \textbf{\ours},  
\textbf{\underline{E}}fficiency-
\textbf{\underline{E}}ffectiveness 
\textbf{\underline{R}}eranking 
\textbf{\textsc{FLOPs}} for LLM-based rerankers: ranking metrics per PetaFLOP (RPP) for relevance per compute and queries per PetaFLOP (QPP) for hardware-agnostic throughput. The proposed metrics thus enable fair comparisons between methods that might utilize different LLMs, reranking algorithms, and running-time choices. 
Accompanied by the proposed metrics, an interpretable FLOPs estimator was built to estimate the FLOPs of an LLM-based reranker even without running any experiments. 

Based on the proposed metrics, we conduct comprehensive experiments to evaluate a wide range of LLM-based rerankers, examining the efficiency-effectiveness trade-off and drawing attention to this issue within the research community. 
% \yilun{move the following sentences to related work Sec 2.2?} 
Our key contributions include:
\begin{itemize}
\item We derive a closed-form, interpretable formula for the FLOPs of LLM-based rerankers and provide an open-source calculator covering up-to-date models and decoding settings.
\item We propose two efficiency-effectiveness metrics: RPP for relevance per compute and QPP for hardware-agnostic throughput.
\item We conduct the first large-scale study of the efficiency–effectiveness trade-off in LLM-based rerankers, bring this issue to the attention of the research community.
\item All code, data, and the FLOPs estimator are publicly released for reproducible research on computationally efficient reranking\metric. 

\end{itemize}

\section{Related Work}

\subsection{LLM-based Rerankers}

Based on how the documents are compared with each other, LLM-based rerankers can be categorized as pointwise, pairwise and listwise.
\textbf{Pointwise} methods primarily compute the query-document relevance score by either the likelihood of generating the query conditioned on the document~\cite{DBLP:journals/sigir/PonteC17, DBLP:conf/sigir/ZhuangZ21a, DBLP:conf/ecir/ZhuangLZ21, DBLP:journals/corr/abs-2404-04522} or the normalized possibility of generating the ``Yes'' when prompting the LLM whether the query-document pair is relevant or not~\cite{DBLP:journals/tmlr/LiangBLTSYZNWKN23, DBLP:conf/emnlp/NogueiraJPL20}. The ranking can be easily accomplished by sorting the relevance score of each document.
\textbf{Pairwise} methods compare the relevance of a pair of documents to a given query and output the document ID of the more relevant one. To rank a list of documents, sorting~\cite{DBLP:conf/naacl/QinJHZWYSLLMWB24} and sampling~\cite{DBLP:conf/ictir/GienappFHP22, DBLP:conf/icpr/MikhailiukW0YM20} methods are proposed. Sorting uses the pairwise comparison to replace the comparison operation in sorting algorithms, such as bubble sorting and heap sorting. In contrast, sampling methods reduce the number of comparisons by repeatedly drawing random pairs (or small subsets), aggregating wins, and estimating a global ranking. Sorting methods are more efficient for getting the top-K documents as they do not need to compare all the pairs. Setwise~\cite{DBLP:conf/sigir/ZhuangZKZ24} extends the pairwise comparison utilized in heapsort and bubblesort to output the best one from three or more documents in one LLM call and thus reduces the number of LLM calls. To rank a list of documents, setwise build \textbf{Listwise} methods directly output a ranked list of document IDs. Most of the existing listwise methods are zero-shot~\cite{DBLP:journals/corr/abs-2305-02156} or few-shot~\cite{DBLP:conf/emnlp/0001YMWRCYR23, DBLP:journals/corr/abs-2305-02156, DBLP:journals/corr/abs-2309-15088} prompting methods. Recently, researchers~\cite{zhang2025rearank} have resorted to adopting reinforcement learning to fine-tune LLMs to generate reasoning followed by ranked document IDs to tackle reasoning-intensive tasks like BRIGHT~\cite{DBLP:conf/iclr/SuYXSMWLSST0YA025}. To get a full rank list, listwise methods usually adopt strategies like sliding-window~\cite{DBLP:conf/emnlp/0001YMWRCYR23} and tournament ranking ~\cite{DBLP:conf/www/0004LZ0MYSMY25} to get a full rank list with a limited window size.\\

\subsection{FLOPs Calculation}
Several FLOPs profilers exist for deep learning models. Still, most are limited to standard forward passes and do not support token-level generation with KV-cache, which is essential for accurate LLM inference estimation. PyPAPI\footnote{\url{https://github.com/flozz/pypapi}} measures CPU-level FLOPs for general Python code but is not designed for PyTorch or GPU workloads. ptflops\footnote{\url{https://github.com/sovrasov/flops-counter.pytorch}} and fvcore\footnote{\url{https://github.com/facebookresearch/fvcore}} compute FLOPs by running a model’s ``forward'' function, but do not support autoregressive decoding. DeepSpeed’s FLOPs profiler~\cite{DBLP:conf/kdd/RasleyRRH20} and calflops\footnote{\url{https://github.com/MrYxJ/calculate-flops.pytorch}} both support the FLOPs of the decoding process, but they also require full forward execution. All existing tools require model execution and lack closed-form support for generation-aware FLOPs estimation. 
For reranking, prior studies utilize coarse FLOP estimates, \eg double the total parameter count ~\cite{DBLP:journals/corr/abs-2504-20595} or open-source tooling ~\cite{abdallah-etal-2025-asrank}, which lack interpretability regarding the specific facts that impact the FLOPs count. Our work differs from theirs in that our FLOPs estimator is well-interpretable, and we propose new metrics to comprehensively evaluate the efficiency and effectiveness of LLM-based rerankers. 

\section{Method}
\label{sec:method}

We first introduce the metrics we designed to measure the efficiency-effectiveness tradeoff of LLM-based rerankers. Then we elaborate the FLOPs estimator that estimates the number of FLOPs needed for one LLM call. To rank a set of documents for a given LLM-based reranker, the number of LLM calls can be estimated, allowing for the estimation of total FLOPs, which can then be compared with those of different LLM-based rerankers.

\subsection{Metrics}

We report two FLOPs-normalized metrics to compare different LLM-based rerankers, thereby effectively evaluating the effectiveness-efficiency tradeoff without being tied to a specific hardware.

\subsubsection{Ranking metrics per PetaFLOP (RPP)}

\begin{equation}
\text{RPP} =
\frac{m(q)}
     {C_\text{q}/10^{15}}
\end{equation}

\noindent where $m(q)$ can be \emph{any} ranking metric for query $q$
(\eg NDCG, MRR, MAP).  
RPP therefore expresses \emph{ranking metrics per peta-FLOP}; a higher value indicates better ranking quality for a fixed compute budget.

\subsubsection{Queries per PetaFLOP (QPP)}

\begin{equation}
\text{QPP} =
\frac{1}{AVG_{C_\text{q}}/10^{15}}
\end{equation}

\noindent QPP measures \emph{throughput}: how many queries can be processed with one
peta-FLOP.  Together, RPP and QPP trace a method’s efficiency–effectiveness
frontier: RPP weights quality per compute, while QPP captures raw
FLOPs-normalized throughput.

\begin{table*}[t!]  
\centering
\small
%\resizebox{0.4\textwidth}{!}{  
\begin{tabular}{l|cc|cc}
\toprule
& \multicolumn{2}{c|}{Multi-head Attention} & \multicolumn{2}{c}{Grouped-query Attention} \\
\midrule
Ops & Parameters & FLOPs per Token & Parameters & FLOPs per Token\\
\midrule
Atten: QKV & $n_{\text{layer}}d_{\text{model}}3d_{\text{attn}}$ & $2n_{\text{layer}}d_{\text{model}}3d_{\text{attn}}$ & $n_{\text{layer}} d_{\text{model}}\,(1 + 2n_{\text{kv}}/n_{q})\,d_{\text{attn}}$ 
  & $2 n_{\text{layer}} d_{\text{model}}\,(1 + 2n_{\text{KV}}/n_{Q})\,d_{\text{attn}}$\\
Atten: O & $n_{\text{layer}}d_{\text{attn}}d_{\text{model}}$ & $2n_{\text{layer}}d_{\text{attn}}d_{\text{model}}$ & $n_{\text{layer}} d_{\text{attn}} d_{\text{model}}$ 
  & $2 n_{\text{layer}} d_{\text{attn}} d_{\text{model}}$ \\
Atten: Mask & - & $4n_{\text{layer}}d_{\text{attn}}d_{\text{model}}$ & -- 
  & $4 n_{\text{layer}} n_{\text{ctx}} (n_{\text{KV}}/n_{Q}) d_{\text{attn}}$ \\
Feedforward & $n_{\text{layer}}2d_{\text{model}}d_{\text{dff}}$ & $2n_{\text{layer}}2d_{\text{model}}d_{\text{dff}}$ & $n_{\text{layer}}\,2 d_{\text{model}} d_{\text{dff}}$ 
  & $2 n_{\text{layer}}\,2 d_{\text{model}} d_{\text{dff}}$\\
\bottomrule
\end{tabular}
 %}
 \caption{FLOP count for the attention mechanism for multi-head attention and grouped-query attention}
\label{tab: attentioncostbigtable}
\end{table*}

\subsection{FLOPs Estimator}

We parameterize a Transformer with four hyper-parameters: the number of layers $n_{\text{layer}}$, the residual-stream width $d_{\text{model}}$, the hidden size of the feed-forward block $d_{\text{ff}}$, and the dimension of attention output $d_{\text{attn}}$ which is the dimension of \texttt{Q}, \texttt{K}, \texttt{V} projections before splitting into multiple heads and by default $d_{\text{attn}}=d_{\text{model}}$. Because decoder-only and encoder–decoder designs dominate LLM-based rerankers, we derive estimates for both. To keep the analysis general yet concrete, we adopt the baseline decoder-only configuration of \citet{DBLP:journals/corr/abs-2001-08361} and the T5 encoder–decoder architecture \citep{DBLP:journals/jmlr/RaffelSRLNMZLL20}. In a typical reranking call, the model receives a prompt (the context, denoted \texttt{ctx}) and produces an output sequence (\texttt{opt}). The prompt concatenates a task-specific prefix $p$, the query $q$, and a list of $w$ documents, resulting in a length of $n_{\text{ctx}}$. The generated sequence has length $n_{\text{opt}}$. 

\subsubsection{Decoder-only}
Following \citet{DBLP:journals/corr/abs-2001-08361}, we ignore sub-leading costs such as nonlinearities, biases, and layer-normalization. The number of attention and feedforward relevant parameters $N_{\text{dec}}$ is:

\begin{equation}\label{eq: n_dec}
\begin{split}
N_{\text{dec}} & \approx 2 d_{\text {model }} n_{\text {layer }}\left(2 d_{\text {attn }}+d_{\text{ff}}\right)
\end{split}
\end{equation}

\noindent Given an LLM with KV cache enabled, we now compute the FLOPs of generating a sequence named \texttt{opt} (short for ``output''), consisting of \( n_{\text{opt}} \) tokens, conditioned on a prompt \texttt{ctx} (short for ``context'') of length \( n_{\text{ctx}} \). The context includes a task-specific prompt \( p \), a query \( q \), and a list of \( w \) documents.
% having $n_{\text{opt}}$ tokens with a prompt \texttt{ctx}, abbreviation of ``context'', the length of which is $n_{\text{ctx}}$, including task-specific prompt $p$, query $q$, and a list of $w$ documents. 
Each token within \texttt{ctx} requires $2 N_{\text{dec}}+4 n_{\text {layer }}n_{\text{ctx}} d_{\text{attn}}$ FLOPs, where $2N_{\text{dec}}$ comes from the fact that each parameter in $N_{\text{dec}}$ has one addition and one multiplication operation and $4 n_{\text {layer }}n_{\text{ctx}} d_{\text{attn}}$ is taken by the basic multi-head attention operation ~\cite{DBLP:journals/corr/abs-2001-08361}. The total FLOPs for $n_{\text{ctx}}$ tokens $C({\text{ctx}})$ is:

\begin{equation}\label{eq: cost_ctx}
C\left(\text{ctx}\right)=2 N_{\text{dec}} n_{\text{ctx}} +4 n_{\text {layer }}n_{\text{ctx}}^{2} d_{\text{attn}}
\end{equation}

\noindent When generating token $\text{opt}_i$, the total sequence length seen by LLM is $n_{\text{ctx}} + (i - 1)$ and the FLOPs for token $i$ is:

\begin{equation}\label{eq: opt_i_flops}
C\left(\text{opt}_i\right)=2 N_{\text{dec}}+4 n_{\text {layer }}\left[n_{\text{ctx}}+(i-1)\right] d_{\text{attn}}
\end{equation}

\noindent The FLOPs of generating $n_{\text{opt}}$ tokens is computed by summing over all the $n_{\text{opt}}$ tokens:

\begin{equation}\label{eq: cost_opt}
\begin{split}
C({\text{opt}}) &= 2 N_{\text{dec}} n_{\text{opt}}\\
&+2 n_{\text {layer }} d_{\text{attn}}\left[2 n_{\text{opt}} n_{\text{ctx}}+n_{\text{opt}}(n_{\text{opt}}-1)\right]
\end{split}
\end{equation}

\noindent The total FLOPs of taking prompt \texttt{ctx} and generating \texttt{opt} is:

\begin{equation}
C(\text{ctx} + \text{opt}) = C(\text{ctx}) + C(\text{opt}) 
\end{equation}

\noindent For LLM-based reranker, $n_{\text{ctx}}$ consists of task-specific prompt $p$, query $q$, and a list of $w$ documents. By approximating the length of each document as the average document length $L_{\text{doc}}$, $n_{\text{ctx}}$ can be estimated as:

\begin{equation}
n_{\text{ctx}} = n_{p} + n_{q} + wl_{\text{doc}}
\end{equation}

\noindent Suppose $n_{\text{Q}}$ represents the number of heads for \texttt{Q} and $n_{\text{KV}}$ denotes the number of heads for \texttt{K} and \texttt{V}. Compared to multi-head attention, the number of parameters and the FLOPs per token changed accordingly, as shown in Table~\ref{tab: attentioncostbigtable}. The equations are rewritten as:

\begin{equation}
\begin{split}
N_{\text{dec}} & \approx 2 d_{\text {model }} n_{\text {layer }}\left((1+\frac{n_{\text{KV}}}{n_{\text{Q}}}) d_{\text {attn }}+d_{\text{ff}}\right)
\end{split}
\end{equation}

\begin{equation}
C\left(\text{ctx}\right)=2 N_{\text{dec}} n_{\text{ctx}} +4 n_{\text {layer }}n_{\text{ctx}}^{2} \frac{n_{\text{KV}}}{n_{\text{Q}}}d_{\text{attn}}
\end{equation}

% \begin{equation}
% \begin{split}
% C({\text{opt}}) &= 2 N_{\text{dec}} n_{\text{opt}}\\
% &+2 n_{\text {layer }} \frac{n_{\text{kv}}}{n_{\text{q}}} d_{\text{attn}}\left[2 n_{\text{opt}} n_{\text{ctx}}+n_{\text{opt}}(n_{\text{opt}}-1)\right]
% \end{split}
% \end{equation}

\begin{equation}
\begin{split}
C({\text{opt}}) =\ & 2 N_{\text{dec}} n_{\text{opt}} + 2 n_{\text{layer}} \frac{n_{\text{KV}}}{n_{\text{Q}}} d_{\text{attn}} \cdot 2 n_{\text{opt}} n_{\text{ctx}} \\
& + 2 n_{\text{layer}} \frac{n_{\text{KV}}}{n_{\text{Q}}} d_{\text{attn}} \cdot n_{\text{opt}}(n_{\text{opt}} - 1)
\end{split}
\end{equation}

For models with MoE, only the parameter count of the ``Feedforward'' component in Table~\ref{tab: attentioncostbigtable} needs to be adjusted. 
Suppose there are $n_{\text{expert}}$ experts, each with intermediate size $d_{\text{dff-MoE}}$. 
Then the number of ``Feedforward'' parameters is 
$n_{\text{layer}} \cdot 2 d_{\text{model}} \cdot n_{\text{expert}} d_{\text{dff-MoE}}$. 
Equivalently, substituting $d_{\text{ff}}$ with $n_{\text{expert}} d_{\text{dff-MoE}}$ yields the updated $N_{\text{dec}}$ for MoE models. 
Because $N_{\text{dec}}$ appears in both $C(\text{ctx})$ and $C(\text{opt})$, these expressions are automatically updated once $N_{\text{dec}}$ is replaced. For instance, Qwen1.5-MoE-A2.7B\footnote{\url{https://huggingface.co/Qwen/Qwen1.5-MoE-A2.7B}} has one expert activated for each token, and for each token, there are four additional experts selected from a pool of 60 experts. Thus, the number of the parameters of ``Feedforward'' is $n_{\text{layer}}2d_{\text{model}}5d_{\text{dff-MoE}}$. Qwen1.5-MoE-A2.7B adopts intermediate size 5632 for each shared expert and 1408 for each of the remaining 60 experts, so $d_{\text{dff-MoE}}=(5632 + 1408*4)/5 = 2252.8$ on average.

\subsubsection{Encoder-Decoder}
Decoder-only LLMs, such as GPT~\cite{radford2019language}, do not include encoder-decoder attention and therefore share a similar structure with the encoder component of encoder-decoder models. The main difference lies in the attention masking strategy, which, however, does not affect the FLOPs required to process the prompt. Although decoder-only models are designed to compute attention only over previous tokens, in practice, they compute full self-attention (\eg $Q_{\text{prompt}} \times K_{\text{prompt}}$) and apply a causal mask to prevent attending to future tokens. So, for encoder-decoder LLMs, the FLOPs of consuming prompt \texttt{ctx} is the same as that of encoder-only LLMs: 

\begin{equation}
C\left(\text{ctx}\right)=2 N_{\text{enc}} n_{\text{ctx}} +4 n_{\text {layer }}n_{\text{ctx}}^{2} d_{\text{attn}}
\end{equation}

\noindent Where $N_{\text{enc}}$ is same as $N_{\text{dec}}$ in Equation~\ref{eq: n_dec}. The decoder employs a different attention mechanism from that of the encoder, utilizing an encoder-decoder attention mechanism followed by self-attention. In an encoder–decoder model, each decoder layer must, once per prompt, project the encoder outputs to cross-attention keys and values. This setup cost is calculated as:
% paid once and then cached
\begin{equation}
C_{\mathrm{cross\text{-}KV}} =
   4\,n_{\text{layer}}\,n_{\text{ctx}}\,d_{\text{model}}\,d_{\text{attn}}
\end{equation}

% Thus, we first revise the $N_{\text{dec}}$ according to Table~\ref{tab: ops_params}. 
\noindent Even, it has two attentions, when generating token $\text{opt}_i$, it only goes through two \texttt{Q} projections, two \texttt{O} one \texttt{K} projection, and one \texttt{V} projection, as the left \texttt{Q} and \texttt{K} projections are for prompt ``ntx'', so

\begin{equation}
\begin{split}
N_{\text{dec}} & \approx 2d_{\text {model }} n_{\text {layer }}\left(3 d_{\text {attn }}+d_{\text{ff}}\right)
\end{split}
\end{equation}

\noindent The total sequence length seen by self-attention is $(i-1)$ and the attention operation takes $4 n_{\text {layer }} (i-1) d_{\text{attn}}$ FLOPs. The sequence length seen by the following encoder-decoder attention is fixed as $n_{\text{ctx}}$, which requires $4 n_{\text {layer }}n_{\text{ctx}} d_{\text{attn}}$ for computing attention. The total attention FLOPs is $4 n_{\text {layer }}(n_{\text{ctx}} + i - 1)d_{\text{attn}}$ which is the same as that of decoder-only models as shown in the right part of equation~\ref{eq: opt_i_flops} and thus the total FLOPs at generating token $\text{opt}_i$ is also same as equation~\ref{eq: opt_i_flops} and the only difference is that the value of $N_{dec}$ is different. Similary, the $C({\text{opt}})$ is same as equation~\ref{eq: cost_opt} but with a different value of $N_{\text{dec}}$. The cost of encoder-decoder model is:

\begin{equation}
C(\text{ctx} + \text{opt}) = C(\text{ctx}) + C_{\mathrm{cross\text{-}KV}} + C(\text{opt}) 
\end{equation}

\newcommand{\rqone}{Can \ours overcome the limitations of existing efficiency proxies?\xspace}
\newcommand{\rqtwo}{What are the LLM-based reranker performances under the \ours?\xspace}
\newcommand{\rqthree}{Do the estimated FLOPs reflect the measured FLOPs?\xspace}
\newcommand{\rqfour}{How does latency relate to the FLOP counts?\xspace}
\newcommand{\rqfive}{What is the relationship between prompt length and FLOPs?\xspace}

\section{Experiment Setup}

Following the setwise setup in~\citet{DBLP:conf/sigir/ZhuangZKZ24}, we utilize the Flan-T5 \cite{DBLP:journals/jmlr/ChungHLZTFL00BW24} as the backbone for most of the LLM-based rerankers except for IRL~\cite{DBLP:conf/iclr/ChenGS25} and Tourrank~\cite{DBLP:conf/www/0004LZ0MYSMY25} as these two methods require a longer context than Flan-T5’s input limitation allows. For IRL and Tourrank, we employ the Llama-3.1-8B-Instruct\footnote{\url{https://huggingface.co/meta-llama/Llama-3.1-8B-Instruct}} model. To compare our estimated FLOPs with those reported by open-source packages, we also include Qwen2.5~\cite{yang2025qwen3} (3B\footnote{\url{https://huggingface.co/Qwen/Qwen2.5-3B-Instruct}}, 7B\footnote{\url{https://huggingface.co/Qwen/Qwen2.5-7B-Instruct}}, 14B\footnote{\url{https://huggingface.co/Qwen/Qwen2.5-14B-Instruct}}), which implements group query attention instead of multi-head attention. For all LLM-based rerankers, we report their performance using the new metrics on the TREC-DL19, TREC-DL20~\cite{craswell2020overview}, and two other datasets from BEIR~\cite{DBLP:journals/corr/abs-2104-08663}. The top 100 documents are retrieved using Pyserini’s BM25~\cite{lin2021pyserini}.

We utilize DeepSpeed’s FLOPs profiler~\cite{DBLP:conf/kdd/RasleyRRH20} and calflops to compute the measured FLOPs and get identical results, so we only report one kind of measured FLOPs. We also present the FLOPs of BM25 in Appendix~\ref{ap: flops_bm25}.

\begin{table*}[t]
	\centering
	\resizebox{1\textwidth}{!}{
		\begin{tabular}{c|l|rrrrrrr|rrrrrrr}
			\toprule
			\multicolumn{2}{c}{} & \multicolumn{7}{|c|}{\textbf{TREC DL 2019}} &  \multicolumn{7}{|c}{\textbf{TREC DL 2020}}
			\\
			\toprule
			&{\small \textbf{Methods}}
			& {\small \textbf{NDCG}}
                & {\small \textbf{\#LLM}}
                & {\small \textbf{In}}
                & {\small \textbf{Out}}
			& {\small \textbf{\#FLOPs}}
			&  {\small \textbf{RPP}}
			& {\small \textbf{QPP}}
			& {\small \textbf{NDCG}}
                & {\small \textbf{\#LLM}}
                & {\small \textbf{In}}
                & {\small \textbf{Out}}
			& {\small \textbf{\#FLOPs}}
			&  {\small \textbf{RPP}}
			& {\small \textbf{QPP}} \\
			\midrule
			
			&
			BM25 &
			.506 & - & - & - & - & - & - &.480& - & - & - & - & - & -\\
			\bottomrule
			\multirow{9}{*}{ \rotatebox[origin=c]{90}{Flan-t5-large}}
			&
			pointwise.qlm &
			.557 & 100 & 152.12 & 0 & 0.009 & 61.89 & \textbf{111.1}* & .567 & 100 & 152.85 & 0 & 0.009 & 63.0 & \textbf{111.1}*\\
			&
			pointwise.yes\_no &
			.654 & 100 & 161.12 & 0 & 0.009 & \textbf{72.67}* & \textbf{111.1}* & .615 & 100 & 161.85 & 0 & 0.009 & \textbf{68.33}* & \textbf{111.1}*\\
			&
			listwise.generation &
			.561& 245 & 486.21 & 10.54 & 0.076 & 7.38 & 13.16 & .547 & 245 & 488.28 & 10.04 & 0.076 & 7.2 & 13.16 \\
			&
			listwise.likelihood &
			.669 & 245 & 384.49 & 0 & 0.058 & 11.53 & 17.24 &\textbf{.626} & 245 & 388.61 & 0 & 0.058 & 10.79 & 17.24\\
                &
			pairwise.allpair & 
			.666 & 9900 & 304.48 & 5 & 1.865 & 0.36 & 0.536 &  .622 & 9900 &  304.47 & 5 & 1.865 & 0.33 & 0.536 \\
			&
			pairwise.heapsort &
			.657 & 230.3 & 455.72 & 10 & 0.066 & 9.95 & 15.15 &  .619 & 226.8 &  459.62 & 10 & 0.066 & 9.38 & 15.15\\
			&
			pairwise.bubblesort &
			.636 & 844.2 & 451.77 & 10 & 0.242 & 2.63 & 4.132 & .589 & 778.5 & 459.03 & 10 & 0.227 & 2.59 & 4.405\\
			&
			setwise.heapsort &
			.670 & 125.4 & 322.65 & 5 & 0.025 & 26.80 & 40.0 & .618 & 124.2 & 325.5 & 5 & 0.025 & 24.72 & 40.0 \\
			&
			setwise.bubblesort &
			\textbf{.678} & 460.5 & 320.90 & 5 & 0.091 & 7.45 & 10.99 & .624 & 457.4 & 325.63 & 5 & 0.092 & 6.78 & 10.87\\
			\bottomrule
			\multirow{9}{*}{ \rotatebox[origin=c]{90}{Flan-t5-xl}}
			& 
			pointwise.qlm &
			.542 & 100 & 152.12 & 0 & 0.034 & 15.94 & \textbf{29.41} & .542 & 100 & 152.85 & 0 & 0.034 & 15.94 & \textbf{29.41}\\
			&
			pointwise.yes\_no &
			.650 & 100 & 161.12 & 0 & 0.036 & \textbf{18.06} & 27.78 &  .636 & 100 & 161.85 & 0 & 0.036 & \textbf{17.67} & 27.78 \\
			&
			listwise.generation &
			.569 & 245 & 486.38  & 11.87 & 0.282 & 2.02 & 3.546 &  .547 & 245 & 489.04 & 11.49 & 0.283 & 1.93 & 3.534 \\
			&
			listwise.likelihood &
			.689 & 245 & 385.49 & 0 & 0.216 & 3.19 & 4.629 &  .672 & 245 & 388.97 & 0 & 0.218 & 3.08 & 4.587\\
                &
			pairwise.allpair &
			\textbf{.713}  & 9900 & 298.33 & 5 & 6.826 & 0.10 & 0.146 &  .682 & 9900 & 297.93 & 5 & 6.817 & 0.10 & 0.147 \\
			&
			pairwise.heapsort &
			.705 & 241.9 & 455.26 & 10 & 0.259 & 2.72 & 3.861 &  \textbf{.692} & 244.3 & 455.76 & 10 & 0.262 & 2.64 & 3.817\\
			&
			pairwise.bubblesort &
			.683 & 886.9 & 451.42 & 10 & 0.942 & 0.73 & 1.061 & .662 & 863.9 & 457.18 & 10 & 0.930 & 0.71 & 1.075 \\
			&
			setwise.heapsort &
			.693 & 129.5 & 321.74 & 5 & 0.096 & 7.22 & 10.42 & .678 & 127.8 & 325.27 & 5 & 0.096 & 7.06 & 10.42\\
			&
			setwise.bubblesort &
			.705 & 446.9 & 335.53 & 5 & 0.346 & 2.04 & 2.890 & .676 & 463.5 & 326.32 & 5 & 0.349 & 1.94 & 2.865 \\
			\bottomrule
			\multirow{9}{*}{ \rotatebox[origin=c]{90}{Flan-t5-xxl}} 
			&
			pointwise.qlm &
			.506 & 100 & 152.12 & 0 & 0.135 & 3.75 & \textbf{7.407} &  .492 & 100 & 152.85 & 0 & 0.136 & 3.62 & \textbf{7.352}\\
			&
			pointwise.yes\_no &
			.644 & 100 & 161.12 & 0 & 0.143 & \textbf{4.50} & 6.993 &  .632 & 100 & 161.85 & 0 & 0.144 & \textbf{4.39} & 6.944\\
			&
			listwise.generation &
			.662 & 245 & 487.08 &  11.53 & 1.105 & 0.60 & 0.904 &  .637 & 245 & 489.60 & 11.05 & 1.110 & 0.57 & 0.901 \\
			&
			listwise.likelihood &
			.701 & 245 & 385.87 & 0 & 0.851 & 0.82 & 1.175 &  .690 & 245 & 389.73 & 0 & 0.860 & 0.8 & 1.162\\
                &
			pairwise.allpair &
			.699 & 9900 & 282.32 & 5 & 25.510 & 0.03 & 0.039 & .688& 9900 & 282.32 & 5 & 25.510 & 0.03 & 0.039\\
			&
			pairwise.heapsort &
			.708 & 239.4 & 456.98 & 10 & 1.010 & 0.70 & 0.990 &  \textbf{.699} & 240.5 & 458.26 & 10 & 1.017 & 0.69 & 0.983\\
			&
			pairwise.bubblesort &
			.679 & 870.5 & 453.06 & 10 & 3.642 & 0.19 & 0.275 &  .681 & 842.9 & 459.56 & 10 & 3.577 & 0.19 & 0.279 \\
			&
			setwise.heapsort &
			.706 & 130.1 & 323.43 & 5 & 0.383 & 1.84 & 2.610 & .688 & 128.1 & 325.01 & 5 & 0.379 & 1.82 & 2.638 \\
			&
			setwise.bubblesort &
			\textbf{.711} & 468.3 & 321.94 & 5 & 1.375 & 0.52 & 0.727 &  .686 & 467.9 & 326.37 & 5 & 1.393 & 0.49 & 0.717 \\
                \bottomrule
			\multirow{2}{*}{ \rotatebox[origin=c]{90}{L3.1}} 
			&
			  IRL &
			.649 & 2 & 4469.12 & 0 & 0.096 & \textbf{6.76} & \textbf{10.42} & .639 & 2 & 4556.31 & 0.0 & 0.098 & \textbf{6.52} & \textbf{10.20} \\
			&
			Tourrank &
			\textbf{.757}* & 130 & 1651.62 & 27.91 & 2.274 & 0.33 & 0.440 &  \textbf{.777} & 130 & 1659.93 & 26.79 & 2.284 & 0.34 & 0.438\\
			\bottomrule
		\end{tabular}
	}
	
\caption{
		Results on TREC DL. All the methods re-rank the top 100 documents retrieved by BM25. \#LLM represents the average number of LLM calls per query for reranking 100 documents. ``In'' and ``Out'' denote the average input tokens and output tokens per LLM call. \#FLOPs is the estimated FLOPs per query for reranking 100 documents. Bold value is the best within each LLM, and starred value is the best across different LLMs. ``L3.1'' represent Llama-3.1-8B-Instruct model. We report NDCG@10 for the NDCG metric.
	}
\label{tab:results}
\end{table*}

\section{Experimental Results and Analysis}
We conduct extensive experiments to study four key research questions. Our results and analysis are as follows:

\paragraph{Q1: \rqone}
Existing efficiency proxies for LLM-based rerankers, such as latency~\cite{DBLP:conf/www/JinPZCZXLFZDM25}, number of LLM calls (\ie, forward passes)~\cite{DBLP:conf/sigir/ZhuangZKZ24}, and token counts~\cite{DBLP:conf/www/0004LZ0MYSMY25}, are weak surrogates for actual compute. \textit{Latency} is confounded by hardware, parallelism, etc.; the same algorithm can appear faster or slower across different platforms. \textit{LLM-call counts} discard model size and sequence length, for instance, one call on a 70B model is orders of magnitude more expensive than one call on a 3B model, yet both count as “1.” \textit{Token counts} ignore model size and the prefill-decode asymmetry (quadratic attention during prefill versus near‑linear growth during decoding), so equal token totals can yield very different FLOPs.

These limitations surface empirically in Table~\ref{tab:results}. With Flan‑T5‑large on DL19, \texttt{pointwise.yes\_no} and \texttt{pointwise.qlm} each use 100 calls but differ in RPP (72.67 vs. 61.89). Fewer calls do not imply proportional gains. On DL19 (Flan-T5-large), \texttt{setwise.heapsort} (125.4 calls) vs \texttt{setwise.bubblesort} (460.5 calls) yields RPP 26.8 vs 7.45. Holding the call count at 100, scaling the backbone from Flan‑T5‑large to ‑xl to ‑xxl collapses RPP/QPP (72.67$\rightarrow$18.06$\rightarrow$4.50; 111.1$\rightarrow$27.78$\rightarrow$6.99), reflecting the sharp FLOPs growth from wider/deeper attention. Similar patterns are observed across different LLMs and datasets (Appendix~\ref{ap: ap_beir})

\begin{figure}[t]
  \centering
  \begin{subfigure}[t]{0.49\linewidth}
    \centering
    \includegraphics[width=\linewidth]{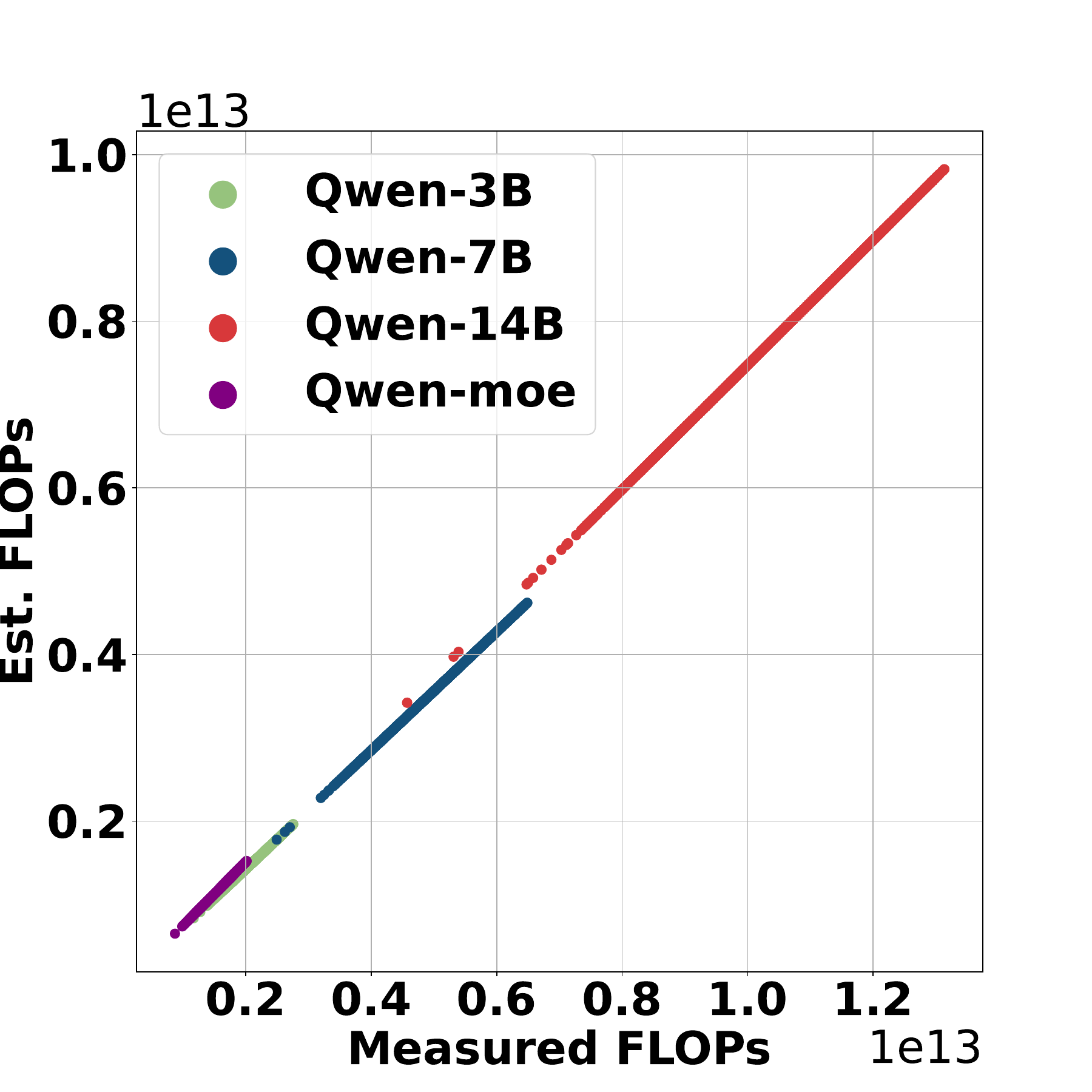}
    \caption{Decoder}
  \end{subfigure}
  \hfill
  \begin{subfigure}[t]{0.49\linewidth}
    \centering
    \includegraphics[width=\linewidth]{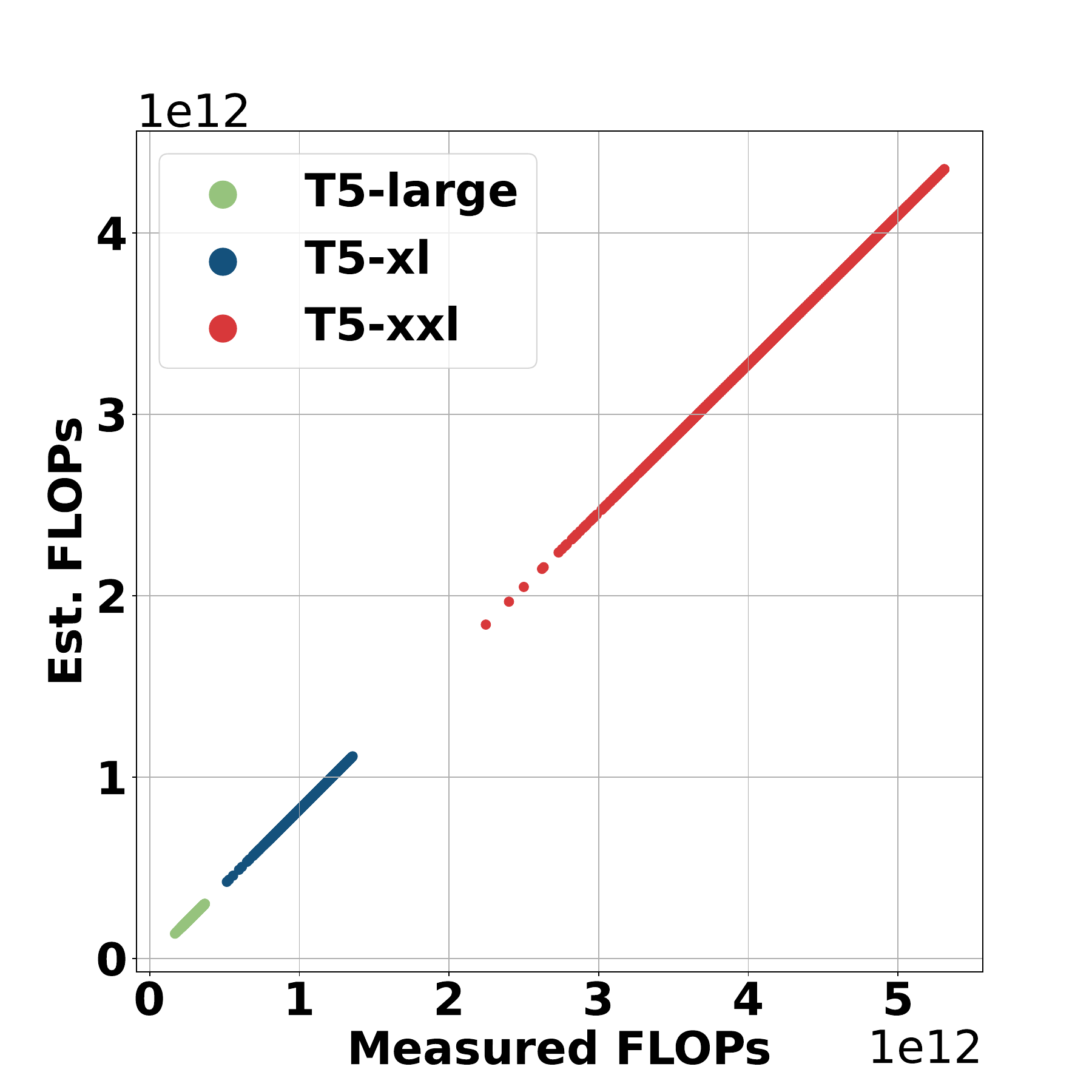}
    \caption{Encoder-Decoder}
  \end{subfigure}
  \caption{Linear trends between estimated and measured FLOPs for decoder (left) and encoder-decoder (right) models of various sizes on DL19. The same is observed for the DL20 dataset.}
  \label{fig:flops}
\end{figure}

\begin{figure}[t]
  \centering
  \begin{subfigure}[t]{0.49\linewidth}
    \centering
    \includegraphics[width=\linewidth]{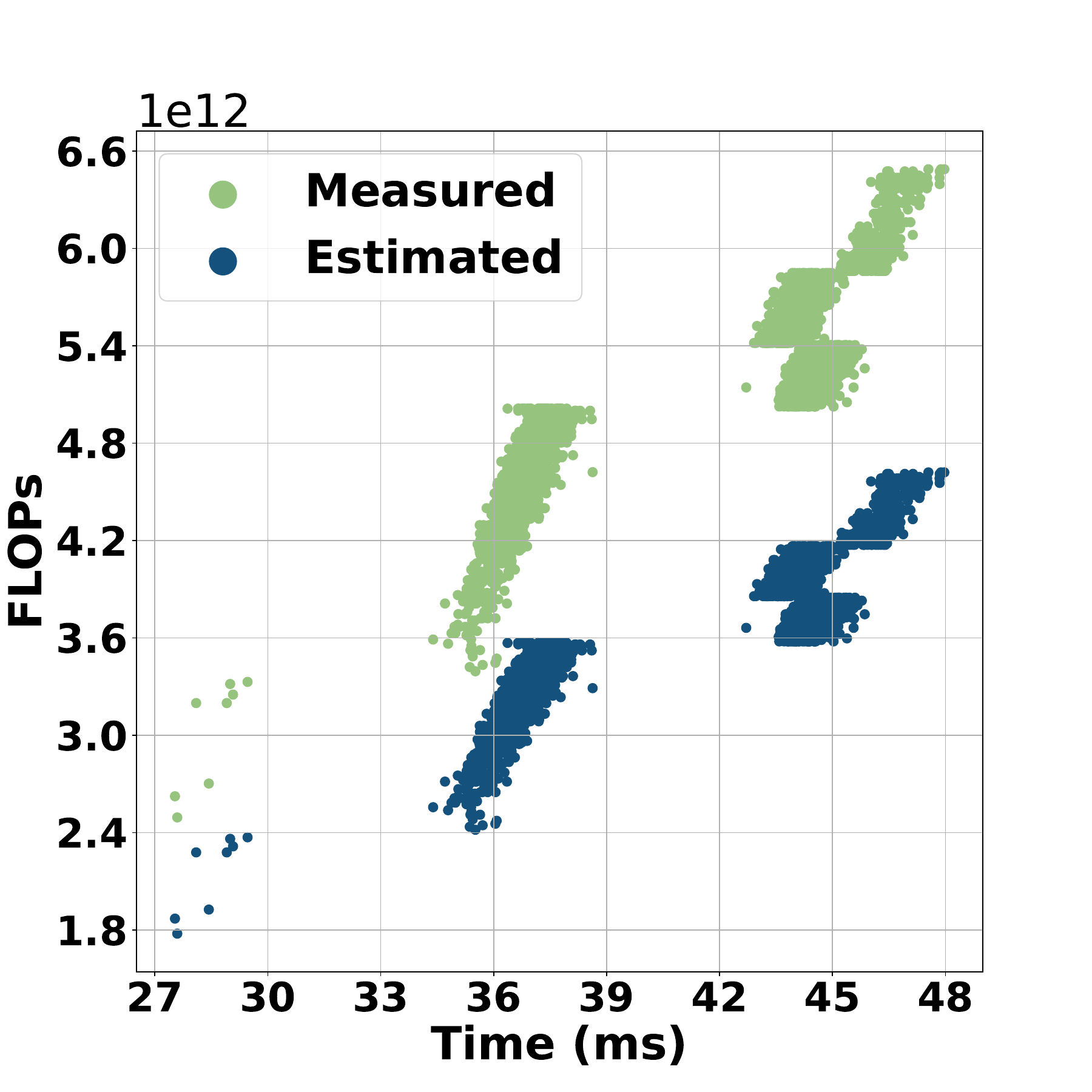}
    \caption{Decoder}
  \end{subfigure}
  \hfill
  \begin{subfigure}[t]{0.49\linewidth}
    \centering
    \includegraphics[width=\linewidth]{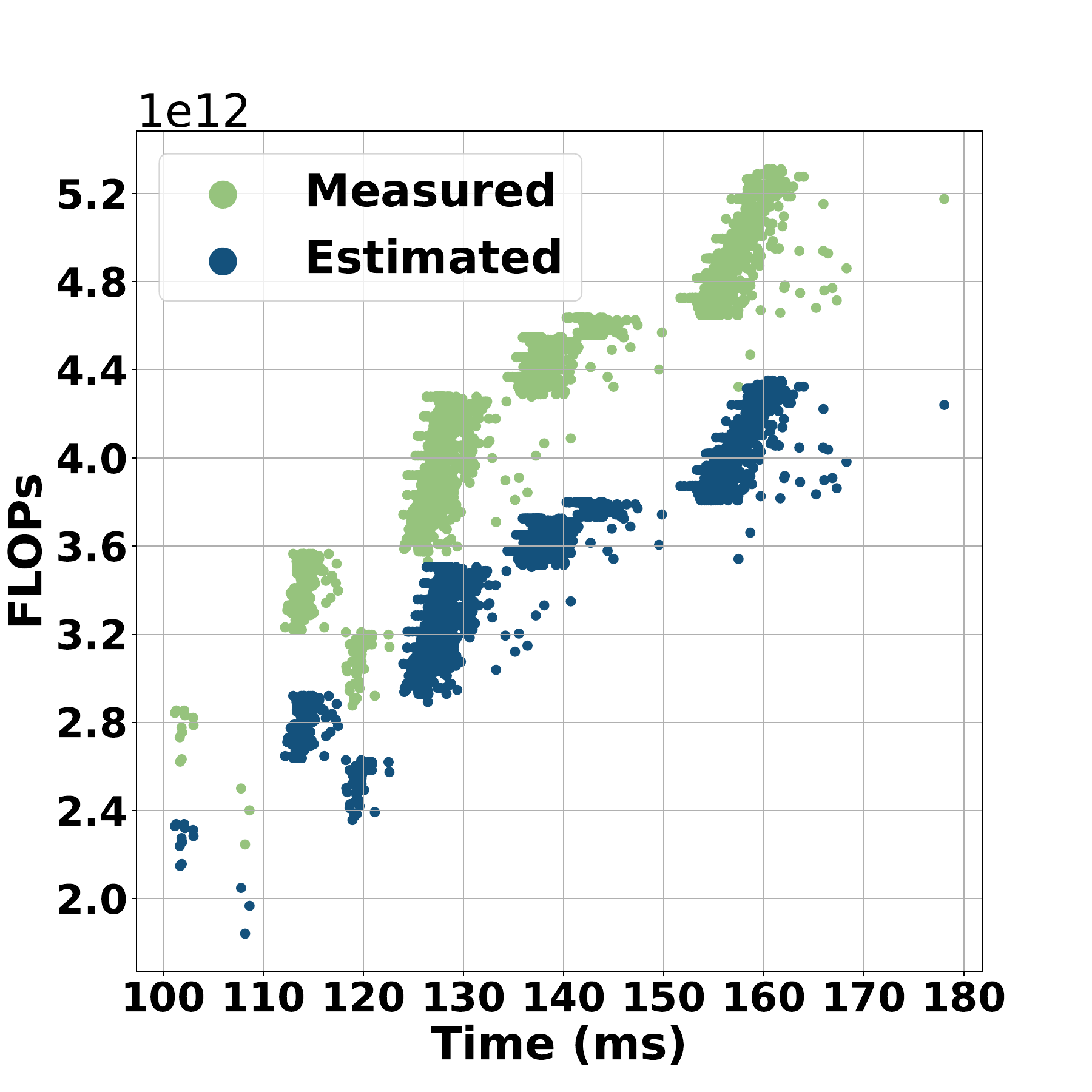}
    \caption{Encoder-Decoder}
  \end{subfigure}
  \caption{Latency in milliseconds increases with FLOPs on Qwen-7B (left) and Flan-T5-XXL (right). The Pearson correlation coefficients between latency and estimated FLOP counts are 0.88 for Qwen-7B and 0.94 for Flan-T5-XXL.}
  \label{fig:latency}
\end{figure}

FLOPs, in contrast, is an intrinsic, hardware‑agnostic measure of work. Normalizing reranking metrics and throughput by FLOPs (RPP/QPP) enables fair, interpretable comparisons across different LLMs. Additionally, it aligns with trends in measured FLOPs and latency (Section \ref{sec:method}, Figure~\ref{fig:flops}, Figure~\ref{fig:latency}).

\paragraph{Q2: \rqtwo}
Table~\ref{tab:results} reports the efficiency–effectiveness trade-offs of a broad set of LLM-based rerankers under the proposed RPP and QPP metrics. Overall, these systems perform poorly once computation is taken into account. Across all LLMs and methods, TourRank with Llama-3.1-8B-Instruct achieves the highest NDCG on both DL19 and DL20, albeit at the cost of almost the lowest RPP and QPP. Beyond this result, several additional insights emerge.

Pointwise methods dominate the RPP and QPP metrics across different LLMs and datasets. \texttt{pointwise.yes\_no} of Flan-T5-large yields the highest RPP of 72.67 (DL19) and 68.3 (DL20) and achieves the maximum QPP of 111 queries/PetoFLOPs. The variant \texttt{pointwise.qlm} obtains a similar QPP metric but 5$\sim$10 worse RPP points. These methods obtain 10\% to 30\% relative NDCG gains over the baseline BM25 with negligible FLOPs consumption compared with other LLM-based rerankers. 

Scaling up hurts efficiency far more than it helps effectiveness. Most LLM-based rerankers gain NDCG when moving from Flan-T5-large to Flan-T5-xl, yet see only marginal improvement from xl to xxl. For example, setwise. Heapsort rises from 0.670 to 0.693 and then to 0.706 in NDCG. Meanwhile, efficiency collapses: RPP plunges from 26.8 to 7.22 and then to 1.84, while QPP falls from 40.0 to 10.42 and finally to 2.61. In short, scaling boosts quality slowly but sacrifices RPP and QPP on a much larger scale.    

Pairwise and listwise methods are intensely FLOP-hungry. Allpair sorting, although it delivers the highest NDCG on Flan-T5-xl (0.713), issues 9,900 LLM calls per query; its RPP collapses to around 0.10, and it processes barely 0.15 queries per petaFLOPs, making large-scale deployment impractical. Heapsort- and bubblesort-based variants cut the number of calls by roughly 90\%, yet remain about orders of magnitude less efficient than pointwise methods on both RPP and QPP. 

\paragraph{Q3: \rqthree}
Figure \ref{fig:flops} shows the relationship between the estimated and measured FLOPs on DL19 for models of various sizes. The comparison contains both decoder-only and encoder-decoder architectures, providing a comprehensive view of scaling trends. In both cases, the estimated and measured FLOP counts scale with the model size, reflecting the expected rise in computational cost with increasing model parameters. 
The linear pattern across models illustrates that the estimated FLOPs correlate linearly with the measured FLOPs and are consistent across model families and architectural types, suggesting that our FLOPs estimator is accurate and reliable. The close alignment between the two quantities provides empirical justification for the FLOPs estimator described in Section~\ref{sec:method}, affirming its reliability as a proxy when real measurements are unavailable.

\paragraph{Q4: \rqfour}
Figure \ref{fig:latency} shows the relationship between latency and FLOP counts for two representative models: Qwen-7B (a decoder-only architecture) and Flan-T5-XXL (an encoder-decoder architecture), evaluated on the DL19 dataset. For both models, we observe that latency increases in accordance with the number of FLOPs. Importantly, the estimated FLOP counts exhibit a correlation with latency that closely mirrors the relationship between measured FLOP counts and latency. The Pearson correlation coefficients between latency and estimated FLOP counts are 0.88 for Qwen-7B and 0.94 for Flan-T5-XXL. This alignment indicates that our FLOPs estimator approximates computational cost accurately and can serve as a reliable predictor of real-world latency trends. This means that the estimator can be used to anticipate inference time without requiring direct hardware profiling, which is particularly useful when comparing models in a platform-agnostic setting or during early-stage architecture design.

\begin{figure}[h]
  \centering
  \begin{subfigure}[t]{0.49\linewidth}
    \centering
    \includegraphics[width=\linewidth]{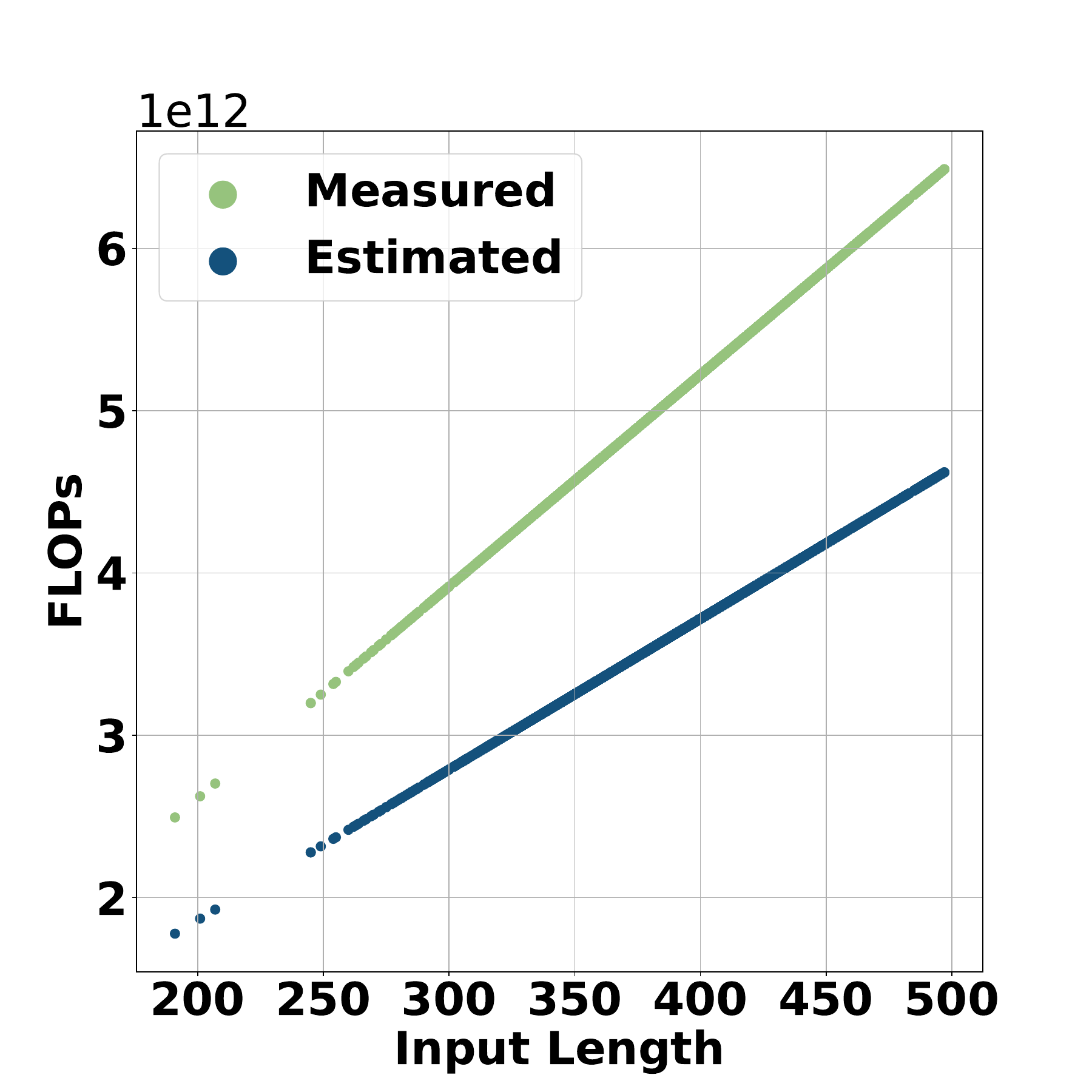}
    \caption{Decoder}
  \end{subfigure}
  \hfill
  \begin{subfigure}[t]{0.49\linewidth}
    \centering
    \includegraphics[width=\linewidth]{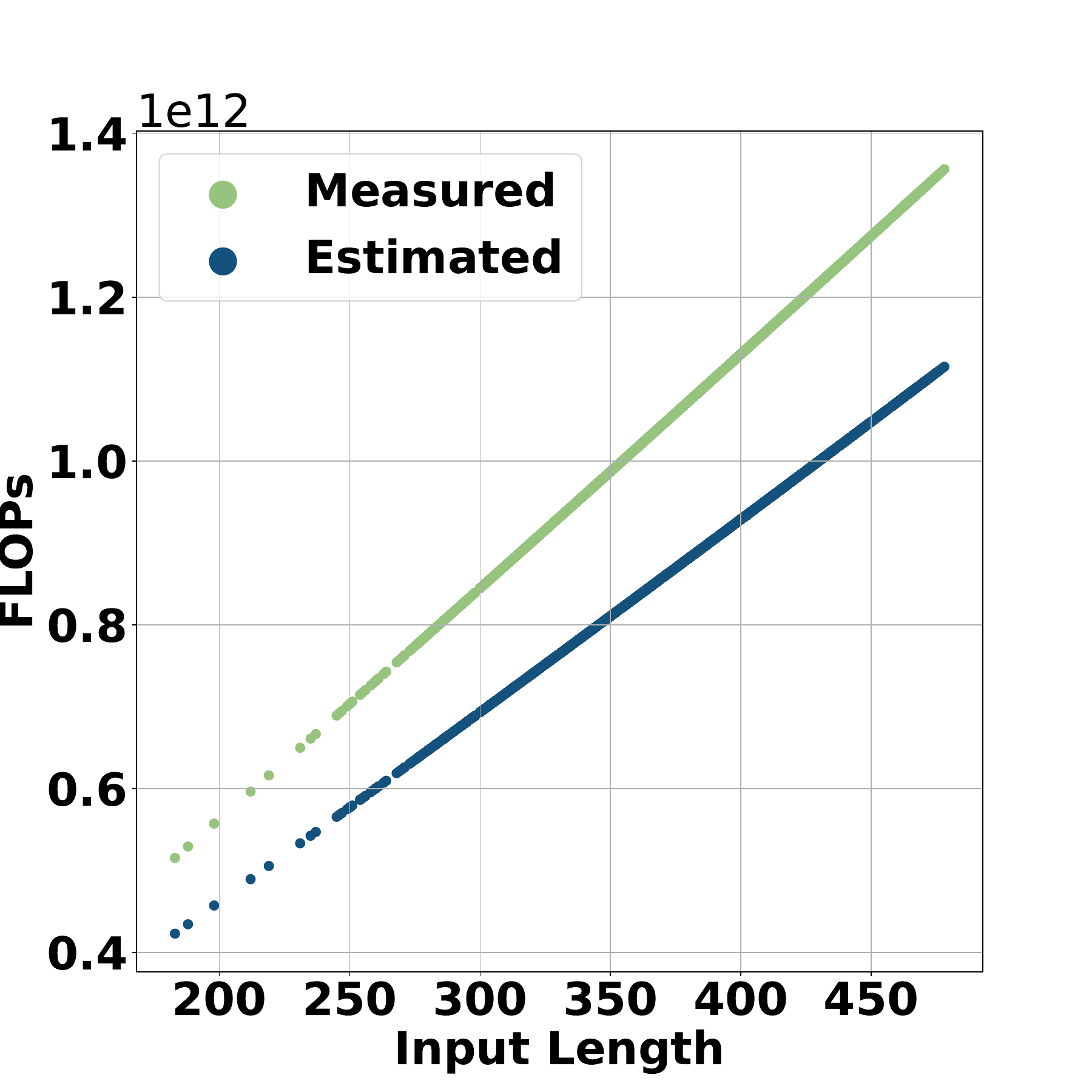}
    \caption{Encoder-Decoder}
  \end{subfigure}
  \caption{FLOPs increases with prompt length for Qwen-7B (left) and Flan-T5-XL (right) on the DL19 dataset.}
  \label{fig:inputlength}
\end{figure}

\paragraph{Q5: \rqfive}
Figure \ref{fig:inputlength} shows the relationship between prompt length and FLOPs. As expected, both the estimated and actual FLOPs increase with longer prompts, reflecting the greater computational cost required to process more input tokens. Notably, the estimated FLOPs exhibit a strong correlation with prompt length, which closely mirrors the pattern observed in the actual FLOPs. The Pearson correlation coefficient between prompt length and FLOP counts is 1. As the prompt becomes longer, the estimator reliably tracks the resulting increase in computation, consistent with what is observed empirically. This result provides additional validation for the robustness of our FLOPs estimator, demonstrating its ability to respond to changes in input length in a manner consistent with measured FLOPs. \\

% ---

% ---

\section{Conclusion and Future Works}

Due to the limitations of existing metrics in evaluating the efficiency-effectiveness tradeoff of large language models as rerankers, we propose two metrics, RPP and QPP, to quantify the model performance. In addition, we provide a calculator based on a closed-form and interpretable formula to compute the FLOPs, and validate this estimation through experiments on existing decoder-only and encoder-decoder model architectures. The estimated FLOP count exhibits a strong linear correlation with the actual measured values, allowing it to approximate real-world computational cost even without model execution. Future work includes conducting a linear regression between the measured and estimated FLOP counts to refine our estimation and adapting to more advanced architectures. 

\clearpage
\newpage
\section*{Limitations}
The FLOP estimation relies on model architecture specifications and assumes consistent implementation across different frameworks, which may not hold in practice due to library-level optimizations or kernel differences. Although the estimator shows strong linear correlation with actual FLOP measurements, the approximation may be less accurate for models with more advanced architectures in the future. While FLOPs offer a more stable proxy than latency or token counts, they do not capture other real-world constraints such as memory bandwidth, energy consumption, or inference-time variability under dynamic system loads.

\bibliography{custom}

\appendix
\begin{table*}[t]
	\centering
	\resizebox{1\textwidth}{!}{
		\begin{tabular}{c|l|rrrrrrr|rrrrrrr}
			\toprule
			\multicolumn{2}{c}{} & \multicolumn{7}{|c|}{\textbf{TREC-COVID}} &  \multicolumn{7}{|c}{\textbf{Robust04}}
			\\
			\toprule
			&{\small \textbf{Methods}}
			& {\small \textbf{NDCG}}
                & {\small \textbf{\#LLM}}
                & {\small \textbf{In}}
                & {\small \textbf{Out}}
			& {\small \textbf{\#FLOPs}}
			&  {\small \textbf{RPP}}
			& {\small \textbf{QPP}}
			& {\small \textbf{NDCG}}
                & {\small \textbf{\#LLM}}
                & {\small \textbf{In}}
                & {\small \textbf{Out}}
			& {\small \textbf{\#FLOPs}}
			&  {\small \textbf{RPP}}
			& {\small \textbf{QPP}} \\
			\midrule
			
			&
			BM25 &
			.595 & - & - & - & - & - & - &.407 & - & - & - & - & - & -\\
			\bottomrule
			\multirow{9}{*}{ \rotatebox[origin=c]{90}{Flan-t5-large}}
			&
			pointwise.qlm & 0.664 & 100 & 160.78 & 0 & 0.010 & 66.40 & 100.00 & 0.439 & 100 & 163.23 & 0 & 0.010 & 43.90 & 100.00 \\
			&
			pointwise.yes\_no & 0.664 & 100 & 169.78 & 0 & 0.010 & 66.40 & 100.00 & 0.456 & 100 & 172.23 & 0 & 0.010 & 45.60 & 100.00 \\ 
			&
			listwise.generation & 0.692 & 245 & 511.55 & 5.15 & 0.080 & 8.65 & 12.50 & 0.441 & 245 & 512.39 & 4.99 & 0.080 & 5.51 & 12.50 \\ 
			&
			listwise.likelihood & 0.756 & 245 & 475.11 & 0 & 0.073 & 10.36 & 13.70 & 0.475 & 245 & 476.64 & 0 & 0.073 & 6.51 & 13.70 \\
                &
			pairwise.heapsort & 0.761 & 241.32 & 630.77 & 10.00 & 0.099 & 7.69 & 10.10 & 0.402 & 182.36 & 634.95 & 10.00 & 0.075 & 5.36 & 13.33 \\ 
			&
			pairwise.bubblesort & 0.714 & 880.18 & 630.84 & 10.00 & 0.361 & 1.98 & 2.77 & 0.439 & 528.48 & 634.06 & 10.00 & 0.218 & 2.01 & 4.59 \\ 
			&
			setwise.heapsort & 0.768 & 129.62 & 449.59 & 5.00 & 0.037 & 20.76 & 27.03 & 0.462 & 120.84 & 452.73 & 5.00 & 0.034 & 13.59 & 29.41 \\ 
			&
			setwise.bubblesort & 0.761 & 468.42 & 450.22 & 5.00 & 0.133 & 5.72 & 7.52 & 0.497 & 462.36 & 452.86 & 5.00 & 0.132 & 3.77 & 7.58 \\
			\bottomrule
			\multirow{9}{*}{ \rotatebox[origin=c]{90}{Flan-t5-xl}}
			& 
			pointwise.qlm & 0.679 & 100 & 160.78 & 0 & 0.036 & 18.86 & 27.78 & 0.427 & 100 & 163.23 & 0 & 0.037 & 11.54 & 27.03 \\ 
			&
			pointwise.yes\_no & 0.698 & 100 & 169.78 & 0 & 0.038 & 18.37 & 26.32 & 0.479 & 100 & 172.23 & 0 & 0.039 & 12.28 & 25.64 \\
			&
			listwise.generation & 0.65 & 245 & 511.44 & 5.08 & 0.293 & 2.22 & 3.41 & 0.475 & 245 & 511.99 & 4.93 & 0.293 & 1.62 & 3.41 \\ 
			&
			listwise.likelihood & 0.736 & 245 & 474.00 & 0 & 0.268 & 2.75 & 3.73 & 0.526 & 245 & 476.40 & 0 & 0.269 & 1.96 & 3.72 \\ 
			&
			pairwise.heapsort & 0.778 & 252.28 & 629.13 & 10.00 & 0.377 & 2.06 & 2.65 & 0.55 & 241.30 & 635.46 & 10.00 & 0.364 & 1.51 & 2.75 \\ 
			&
			pairwise.bubblesort & 0.763 & 914.34 & 628.55 & 10.00 & 1.365 & 0.56 & 0.73 & 0.553 & 771.66 & 634.93 & 10.00 & 1.164 & 0.48 & 0.86 \\ 
			&
			setwise.heapsort & 0.757 & 135.8 & 449.53 & 5.00 & 0.142 & 5.33 & 7.04 & 0.52 & 129.38 & 452.73 & 5.00 & 0.136 & 3.82 & 7.35 \\
			&
			setwise.bubblesort & 0.756 & 468.96 & 449.38 & 5.00 & 0.491 & 1.54 & 2.04 & 0.537 & 452.44 & 452.86 & 5.00 & 0.477 & 1.13 & 2.10 \\ 
			\bottomrule
			\multirow{9}{*}{ \rotatebox[origin=c]{90}{Flan-t5-xxl}} 
			&
			pointwise.qlm & 0.707 & 100 & 160.78 & 0 & 0.143 & 4.94 & 6.99 & 0.44 & 100 & 163.23 & 0 & 0.146 & 3.01 & 6.85 \\ 
			&
			pointwise.yes\_no & 0.691 & 100 & 169.78 & 0 & 0.152 & 4.55 & 6.58 & 0.515 & 100 & 172.23 & 0 & 0.154 & 3.34 & 6.49 \\ 
			&
			listwise.generation & 0.664 & 245 & 511.44 & 5.19 & 1.147 & 0.58 & 0.87 & 0.495 & 245 & 511.98 & 5.01 & 1.148 & 0.43 & 0.87 \\ 
			&
			listwise.likelihood & 0.749 & 245 & 474.73 & 0 & 1.052 & 0.71 & 0.95 & 0.518 & 245 & 476.43 & 0 & 1.056 & 0.49 & 0.95 \\ 
                &
			
			pairwise.heapsort & 0.738 & 243.08 & 629.60 & 10.00 & 1.416 & 0.52 & 0.71 & 0.543 & 244 & 635.42 & 10.00 & 1.434 & 0.38 & 0.70 \\ 
			&
			 pairwise.bubblesort & 0.733 & 887.98 & 630.01 & 10.00 & 5.175 & 0.14 & 0.19 & 0.55 & 859.26 & 635.05 & 10.00 & 5.049 & 0.11 & 0.20 \\ 
			&
			setwise.heapsort & 0.752 & 135.66 & 449.48 & 5.00 & 0.557 & 1.35 & 1.80 & 0.513 & 132.12 & 452.72 & 5.00 & 0.546 & 0.94 & 1.83 \\ 
			&
			 setwise.bubblesort & 0.768 & 470.88 & 449.89 & 5.00 & 1.935 & 0.40 & 0.52 & 0.534 & 456.23		 & 452.86 & 5.00 & 1.887 & 0.28 & 0.53 \\
                \bottomrule
		\end{tabular}
	}
\caption{
		Results on TREC-COVID and Robust04. All the methods re-rank BM25 top 100 documents. \#LLM represents average number of LLM call per query for reranking 100 documents. ``In'' and ``Out'' denote the average input tokens and output tokens per LLM call. \#FLOPs is the estimated FLOPs per query for reranking 100 documents. Bold value is the best within each LLM and stared value is the best across different LLMs. ``L3.1'' represent Llama-3.1-8B-Instruct model. We report NDCG@10 for the NDCG metric.
	}
\label{tab:results-more}
\end{table*}
\section{Appendix}
\label{sec:appendix}

\subsection{BM25}\label{ap: flops_bm25}

\begin{equation}
\sum_{i=1}^{n} \text{IDF}(q_i) \frac{f(q_i, D) (k + 1)}{f(q_i, D) + k \left(1 - b + b \frac{|D|}{\text{avgdl}}\right)}
\label{eq: bm25}
\end{equation}

BM25 \cite{robertson2009probabilistic} computes a relevance score between a query and a document with Equation \ref{eq: bm25} where $\text{IDF}$ denotes the inverse document frequency, $f(q_i, D)$ represents the frequency of the $i^{th}$ query token in the document, $|D|$ is the length of the document, $\text{avgdl}$ is the average document length in the corpus, and $k$ and $b$ are hyperparameters. Assuming that the term frequencies and the inverse document frequencies are precomputed and do not contribute to the runtime FLOP count, the upper bound on the FLOPs required for BM25 scoring can be estimated as:

\begin{equation}
C(\text{BM25}) = 11 \cdot L_Q \cdot N_D
\end{equation}

\noindent where $L_Q$ is the length of the query and $N_D$ is the number of documents, which is 100 for reranking top-100 documents in our experiments. This represents an upper bound because it assumes that every query token appears in every document. In cases where a query token does not appear in a document, the corresponding numerator becomes zero, resulting in zero FLOPs for that term-document pair instead of 11.

\subsection{BEIR results}\label{ap: ap_beir}
We conduct extra experiments on two more datasets from BEIR: TREC-COVID and Robust04 to strengthen generalizability. The results are shown in Table~\ref{tab:results-more}. We observe the same efficiency–effectiveness patterns as in Table~\ref{tab:results}. Pointwise rerankers consistently deliver the strongest RPP and the highest QPP, whereas stronger models get the better performance at the cost of RPP and QPP. On TREC-COVID (Flan-T5-large), \texttt{setwise.heapsort} vs.\ \texttt{setwise.bubblesort} yields RPP 20.76 vs.\ 5.72 (QPP 27.03 vs.\ 7.52); on Robust04 the gap is 13.59 vs.\ 3.77 (29.41 vs.\ 7.58). Scaling the backbone collapses FLOPs-normalized efficiency: \texttt{pointwise.yes\_no} QPP falls 100.00$\rightarrow$26.32$\rightarrow$6.58 on TREC-COVID (RPP 66.40$\rightarrow$18.37$\rightarrow$4.55) and 100.00$\rightarrow$25.64$\rightarrow$6.49 on Robust04 (RPP 45.60$\rightarrow$12.28$\rightarrow$3.34), while NDCG gains remain modest. For both Table~\ref{tab:results} and Table~\ref{tab:results-more}, \texttt{setwise.heapsort} achieves top-tier performance while maintaining good efficiency.

\end{document}